\documentclass[11pt,a4paper]{article}
\usepackage[hyperref]{acl2021}
\usepackage{times}
\usepackage{latexsym}

\usepackage{graphicx}
\usepackage{url}
\usepackage{xspace}
\graphicspath{ {images/} }

\usepackage{alphabeta}

\usepackage{microtype}

\aclfinalcopy 
\def\aclpaperid{***}

\newcommand\dataset{\textsc{ccc}\xspace}
\newcommand\oc{\textsc{oc}\xspace}
\newcommand\ic{\textsc{ic}\xspace}

\title{Toxicity Detection can be Sensitive to the Conversational Context}

\author{Alexandros Xenos$^\clubsuit$, John Pavlopoulos$^\spadesuit$$^\clubsuit$\thanks{~~Corresponding author.}, Ion Androutsopoulos$^\clubsuit$,\\
\textbf{Lucas Dixon$^\circ$, Jeffrey Sorensen$^\ddagger$, Léo Laugier$^\diamond$}\\
$^\clubsuit${Department of Informatics, Athens University of Economics and Business, Greece}\\
$^\spadesuit${Department of Computer and Systems Sciences, Stockholm University, Sweden}\\
$^\diamond$ {Télécom Paris, Institut Polytechnique de Paris, France}\\
$^\ddagger$ Google Jigsaw\\
$^\circ$ Google\\

{\tt \{a.xenos20,annis,ion\}@aueb.gr}\\
{\tt leo.laugier@telecom-paris.fr}\\
{\tt \{ldixon,sorenj\}@google.com}
}
\date{}

\begin{document}
\maketitle
\begin{abstract}
\vspace*{-1mm}
User posts whose perceived toxicity depends on the conversational context are rare in current toxicity detection datasets. Hence, toxicity detectors trained on existing datasets will also tend to disregard context, making the detection of context-sensitive toxicity harder when it does occur. We construct and publicly release a dataset of 10,000 posts with two kinds of toxicity labels: (i) annotators considered each post with the previous one as context; and (ii) annotators had no additional context. Based on this, we introduce a new task, \emph{context sensitivity estimation}, which aims to identify posts whose perceived toxicity changes if the context (previous post) is also considered. We then evaluate machine learning systems on this task, showing that classifiers of practical quality can be developed, and we show that data augmentation with knowledge distillation can improve the performance further.
Such systems could be used to enhance toxicity detection datasets with more context-dependent posts, or to suggest when moderators should consider the parent posts, which often may be unnecessary and may otherwise introduce significant additional cost.
\end{abstract}

\section{Introduction}
Online fora are used to facilitate discussions, but can suffer from hateful, insulting, identity-attacking, profane, or otherwise abusive posts. Such posts are called toxic \cite{Borkan2019}, offensive \cite{Sap2020} or abusive \cite{Thylstrup2020}, and systems detecting them \cite{waseem-hovy-2016-hateful,pavlopoulos-etal-2017-deeper,Badjatiya} are called toxicity (or offensive or abusive language) detection systems. 
What most of these systems have in common, besides aiming to promote healthy discussions online \cite{zhang-etal-2018-conversations}, is that they disregard much of the conversational context, making the detection of context-sensitive toxicity a lot harder.

We consider \textit{context} to be information relevant to help understand the 
meaning and intention of a post; when context is missing, there is more ambiguity in the interpretation of a post.
Context is very diverse in nature, because human communication is diverse; people may inhabit any number of roles in their relationships with 
others.
A person on stage in a play about a murder might engage in dialog that would be illegal in other contexts.
Far from being inappropriate, people may pay to see this behavior and applaud it.
It is not always clear what social norms, jurisdictional mandates, and enforcement regimes apply. A comedian may
deliberately engage in provocative language to entertain, inspire or critique society, but a disruptive heckler might still be removed
by the venue's bouncers.

In online discourse, context typically includes personal information about the authors \cite{pavlopoulos-etal-2017-improved}, the interlocutors, metadata about the conversation or subtle references to specific subjects and topics. 
Within the scope of this work we presume some socially constructed context in the form of common notions about
what constitutes appropriate communicative intent in a social media setting 
-- at least enough that
persons tasked with evaluating the communicative intent can consensually make judgements from the surface text alone. 
This concept, of 
common socially agreed norms, is obviously not a black and white concept, and while certainly worthy of deeper analysis, it is not the focus of our study here. Instead we follow the common practice of having these background social norms 
manifested through crowd-sourcing platforms and measured at a very abstract level by inter-annotator agreement metrics. 
Given this approach, we focus on the past conversational context, specifically the previous post in a discussion. 
For instance, a post ``Keep the hell out'' 
is likely to be considered as toxic by a moderator who has not seen that the previous post was ``What was the title of that `hell out' movie?''.

Although toxicity datasets that include conversational context have recently started to appear, in previous work we showed that context-sensitive posts seem to be rare and this makes it hard for models to learn to detect context-dependent toxicity~\cite{pavlopoulos2020toxicity}. To study this problem, we construct and publicly release a context-aware dataset of 10,000 posts, each of which was annotated by raters who (i) considered the previous (\emph{parent}) post as context, apart from the post being annotated (the \emph{target} post), and by raters who (ii) were given only the target post, without any
other previous conversational context.\footnote{The dataset is released under a CC0 licence.
It can be downloaded from \url{https://github.com/ipavlopoulos/context_toxicity/tree/master/data}.
}

We limit the conversational context to the previous post of the thread, 
as in our previous work \cite{pavlopoulos2020toxicity}, as a first step towards studying broader context-dependent toxicity. While this is still a very limited form of context, our previous work also highlighted the basic challenges of studying context: it is expensive and time consuming to consider on crowd-sourcing platforms, because of the challenges of ensuring that a person has in fact considered the context. The more context, and more subtle kinds of context, one attempts to include in a study, the harder it is to ensure annotators have accounted for it. Moreover context sensitive toxicity in posts is also rare; and thus it is reasonable to wonder if the impact of more indirect and subtle kinds of context is rarer still.

We then use our new dataset to study the nature of context sensitivity in toxicity detection, and we introduce a new task, \emph{context sensitivity estimation}, which aims to identify posts whose perceived toxicity changes if the context (previous post) is also considered. Using the dataset, we also show that systems of practical quality can be developed for the new task. Such systems could be used to enhance toxicity detection datasets with more context-dependent posts, or to suggest when moderators should consider the parent posts, which may not always be necessary and may otherwise introduce significant additional cost. Finally, we show that data augmentation with teacher-student knowledge distillation can further improve the performance of context sensitivity estimators.

\section{The New Dataset (CCC)}
\label{The_dataset} \label{sec:CCC}

To build the dataset of this work, we used the publicly available Civil Comments (CC) dataset \cite{Borkan2019}. CC was originally annotated by ten annotators per post, but the parent post (the previous post in the thread) was not shown to the annotators. We randomly sampled 10,000 CC posts and gave both the target and the parent post to the annotators. We call this new dataset Civil Comments in Context (CCC). Each CCC post was rated either as \textsc{non-toxic}, \textsc{unsure}, \textsc{toxic}, or \textsc{very toxic}, as in the original CC dataset. We unified the latter two labels in both CC and CCC to simplify the problem. To obtain the new in-context labels of CCC, we used the APPEN platform and five high accuracy annotators per post (annotators from zone 3, allowing adult and warned for explicit content),\footnote{\url{https://appen.com}} selected from 7 English speaking countries, namely: UK, Ireland, USA, Canada, New Zealand, South Africa, and Australia.\footnote{We chose populous majority  English-speaking countries. The most common country of origin was USA.}

\begin{figure}[tb]
\centering
\includegraphics[width=0.3\textwidth]{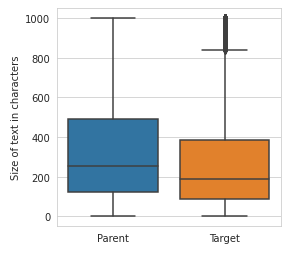}
\captionof{figure}{Length of parent/target posts in characters. }
\label{fig:data_size}
\end{figure}

The free-marginal kappa, a measure of inter-annotator agreement \cite{Randolph}, of the CCC annotations is 83.93 percent, while the average (mean pairwise) percentage agreement is 92 percent. In only 71 posts (0.07 percent) an annotator said \textsc{unsure}, meaning annotators were confident in their decisions most of the time. We exclude these 71 posts from our study, as there are too few to generalize about. The average length of target posts in CCC is only slightly lower than that of parent posts. Figure ~\ref{fig:data_size} shows this when counting the length in characters, but the same holds when counting words (56.5 vs.\ 68.8 words on average). To obtain a single toxicity score per post, we calculated the percentage of the annotators who found the post to be insulting, profane, identity-attack, hateful, or toxic in another way; all toxicity sub-types provided by the annotators were collapsed to a single toxicity label. This is similar to arrangements in the work of \citet{10.1145/3038912.3052591}, who also found that training using the empirical distribution (over annotators) of the toxic labels (a continuous score per post) leads to better toxicity detection performance, compared to using labels reflecting the majority opinion of the raters (a binary label per post). See also \citet{fornaciari-etal-2021-beyond}.

\begin{figure}[h]
\centering
\includegraphics[width=.9\columnwidth]{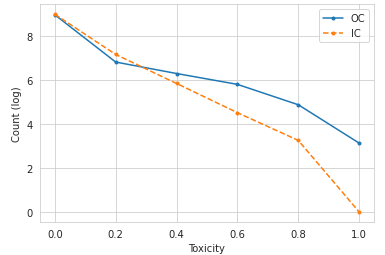}
\caption{Histogram (converted to curve) showing the distribution of toxicity scores according to annotators who were (\ic) or were not (\oc) given the parent posts.
}
\label{fig:toxicity_oc_vs_toxicity_ic}
\end{figure}

Combined with the original (out of context) annotations of the 10,000 posts from CC, the new dataset (CCC) contains 10,000 posts for which both in-context (\ic) and out-of-context (\oc) labels are available. 
Figure \ref{fig:toxicity_oc_vs_toxicity_ic} shows the number of posts (Y axis) per ground truth toxicity score (X axis). Orange (dashed) represents the ground truth obtained by annotators who were provided with the parent post when rating (\ic), while blue (solid) is for annotators who rated the post without context (\oc). The vast majority of the posts were unanimously perceived as \textsc{non-toxic} (0.0 toxicity), both by the \oc and the \ic coders. However, \ic coders found fewer posts with toxicity greater than 0.2, compared to \oc coders. This is consistent with the findings of our previous work \cite{pavlopoulos2020toxicity}, where we observed that when the parent post is provided, the majority of the annotators perceive fewer posts as toxic, compared to showing no context to the annotators. To study this further, in this work we compared the two annotation scores (\ic, \oc) per post, as discussed below.

\begin{figure}[tb]
\includegraphics[scale=0.7]{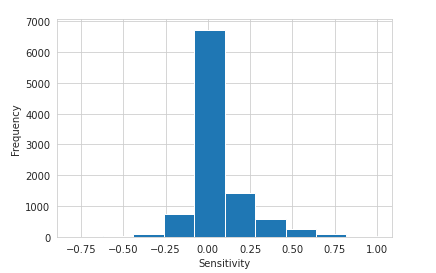}
\caption{Histogram of context sensitivity. Negative (positive) sensitivity means the toxicity increased (decreased) when context was shown to the annotators.}
\label{fig:hist_context_sensitivity}
\end{figure}

For each post $p$, we define $s^{ic}(p)$ to be the toxicity score (fraction of coders who perceived the post as toxic) derived from the \ic coders, and $s^{oc}(p)$ to be the toxicity derived from the \oc coders. Then, their difference is $δ(p) =s^{oc}(p)-s^{ic}(p)$. A positive $δ$ means that raters who were not given the parent post perceived the target post as toxic more often than raters who were given the parent post. A negative $δ$ means the opposite. Figure\ \ref{fig:hist_context_sensitivity} shows that $δ$ is most often 0,  but when the toxicity score changes, $δ$ is most often positive, i.e., showing the context to the annotators reduces the perceived toxicity in most cases. In numbers, in 66.1 percent of the posts the toxicity score remained unchanged while out of the remaining 33.9 percent, in 9.6 percent it increased (960 posts) and in 24.3 percent it decreased (2,408) when context was provided. If we binarize the ground truth (both for \ic and \oc) we get a similar trend, but with the toxicity of more posts remaining unchanged (i.e., 94.7 percent).

\begin{figure}[h]
\includegraphics[width=0.45\textwidth]{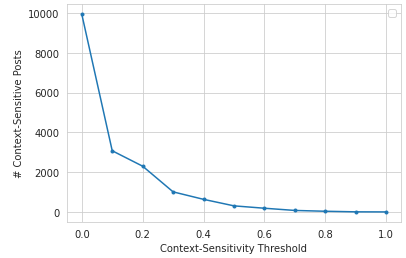}
\caption{Number of context-sensitive posts ($|\delta| \geq t$), when varying the context-sensitivity threshold $t$.}
\label{fig:context_sensitive_posts_per_delta}
\end{figure}

When counting the number of posts for which $|δ|$ exceeds a threshold $t$, called \emph{context-sensitive posts} in Figure~\ref{fig:context_sensitive_posts_per_delta}, we observe that as $t$ increases, the number of context sensitive posts decreases. This means that clearly context sensitive posts (e.g., in an edge case, ones that all \oc coders found as toxic while all \ic coders found as non toxic) are rare. Some examples of target posts, along with their parent posts and $\delta$, are shown in Table \ref{table:examples}.

\begin{table*}[h]
{\small
\centering
\begin{tabular}{ |p{5.5cm}|p{5.5cm}|c|c|c|c| } 
\hline
&&&&
\\
 \sc parent of post $p$ & \sc post $p$ &  $s^{OC}(p)$ & $s^{IC}(p)$ & $\delta$\\
\hline
 Oh Don..... you are soooo predictable. & oh Chuckie you are such a tattle tale. & 36.6\% & 80\% & -43.4\%\\ 
 \hline
 Oh Why would you wish them well? They've destroyed the environment in their country and now they are coming here to do the same. & ``They"?  Who is they?  Do all Chinese look alike to you?  Or are you just revealing your innate bigotry and racism? & 70\% & 0\% & 70\% \\ 
 \hline
\end{tabular}
}
\caption{Examples of context-sensitive posts in \dataset. Here $s^{\oc}(p)$ and $s^{\ic}(p)$ are the fractions of out-of-context or in-context annotators, respectively, who found the target post $p$ to be toxic; and $\delta = s^{\oc}(p)-s^{\ic}(p)$.}
\label{table:examples}
\end{table*}

\section{Experimental Study}
\label{Experimental_Study}

Initially, we used our dataset to experiment with existing \emph{toxicity detection} systems, aiming to investigate if context-sensitive posts are more difficult to automatically classify correctly as toxic or non-toxic. Then, we trained new systems to solve a different task, that of estimating how sensitive the toxicity score of each post is to its parent post, i.e., to estimate the
\emph{context sensitivity} of a target post.

\subsection{Toxicity Detection}
\begin{figure}[h]
\centering
\includegraphics[width=0.48\textwidth]{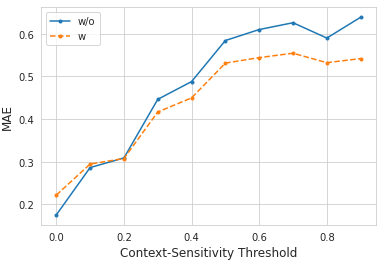}
\caption{Mean Absolute Error (Y-axis) when \emph{predicting toxicity} for different context-sensitivity thresholds ($t$; X-axis). We applied Perspective to target posts alone (w/o) or concatenating the parent posts (w).}
\label{fig:Perspective_exps}
\end{figure}

We employed the Perspective API toxicity detection system, as is and with no further fine-tuning, to classify CCC posts as toxic or not.\footnote{\url{https://www.perspectiveapi.com}} We either concatenate the parent post to the target one to allow the model to “see” the parent, or not.\footnote{We are investigating better context-aware models.} Figure~\ref{fig:Perspective_exps} shows the Mean Absolute Error (MAE) of Perspective, with and without the parent post concatenated, when evaluating on all the CCC posts ($t=0$) and when evaluating on smaller subsets with increasingly context-sensitive posts ($|\delta| \geq t$,  $t>0$). In all cases, we use the in-context (\ic) gold labels as the ground truth. The greater the sensitivity threshold $t$, the smaller the sample (Figure~\ref{fig:context_sensitive_posts_per_delta}).

Figure~\ref{fig:Perspective_exps} shows that
when we concatenate the parent to the target post (w curve), MAE is clearly smaller, provided that $t\geq0.2$. Hence, the benefits of integrating context in toxicity detection systems may be visible only in sufficiently context-sensitive subsets, like the ones we would obtain by evaluating (and training) on posts with $t \geq 0.2$. By contrast, if no context-sensitivity threshold is imposed ($t=0$) when constructing a dataset, the non-context sensitive posts ($|\delta| =0$) dominate (Figure~\ref{fig:context_sensitive_posts_per_delta}), hence adding context mechanisms to toxicity detectors has no visible effect in test scores. This explains related observations in our previous work \cite{pavlopoulos2020toxicity}, where we found that context-sensitive posts are too rare and, thus, context-aware models do not perform better on existing toxicity datasets.

Notice that the more we move to the right of Figure~\ref{fig:Perspective_exps}, the higher the error for both Perspective variants (with, without context). This is probably due to the fact that Perspective is trained on posts that have been rated by annotators who were not provided with the parent post (out of context; \oc), whereas here we use the in-context (\ic) annotations as ground truth. The greater the $t$ in Figure~\ref{fig:Perspective_exps}, the larger the difference between the toxicity scores of \oc and \ic annotators, hence the larger the difference between the (\oc) ground truth that Perspective saw during its training and the ground truth that we use here (\ic). Experimenting with artificial parent posts (long or short, toxic or not) confirmed that the error increases for context-sensitive posts.

The  solution  to  the  problem of increasing error as context sensitivity increases (Figure~\ref{fig:Perspective_exps}) would be to train toxicity detectors on datasets that are richer in context-sensitive posts. However, as such posts are rare (Figure \ref{fig:context_sensitive_posts_per_delta}) they are hard to collect and annotate. This observation motivated the experiments of the next section, where we train \mbox{\textbf{\textit{context-sensitivity}}} detectors, which allow us to collect posts that are likely to be context-sensitive. These posts can then be used to train toxicity detectors on datasets richer in context-sensitive posts.

\subsection{Context Sensitivity Estimation} 
\label{context_sensitivity_experiments}
We trained and assessed four regressors on the new CCC dataset, to predict the context-sensitivity $δ$. We used Linear Regression, Support Vector Regression, a Random Forest regressor, and a BERT-based \cite{devlin-etal-2019-bert} regression model (BERTr). The first three regressors use TF-IDF features. In the case of BERTr, we add a feed-forward neural network  (FFNN) on top of the top-level embedding of the [CLS] token. The FFNN consists of a  dense layer (128 neurons) and a \textit{tanh} activation function, followed by another dense layer. The last dense layer has a single output neuron, with no activation function, that produces the context sensitivity score. Preliminary experiments showed that adding simplistic context-mechanisms (e.g., concatenating the parent post) to the context sensitivity regressors does not lead to improvements. This may be due to the fact that it is often possible to decide if a post is \emph{context-sensitive} or not (we do not score the toxicity of posts in this section) by considering only the target post without its parent (e.g., in responses like “NO!!”). Future work will investigate this hypothesis further by experimenting with more elaborate context-mechanisms. If the hypothesis is verified, manually annotating context-sensitivity (not toxicity) may also require only the target post.  

\begin{table}[tb]
\centering
\small
\begin{tabular}{ |c|c|c|c|c|c|c|} 
\hline
  & \sc MSE $\downarrow$ & \sc MAE $\downarrow$  & \sc AUPR $\uparrow$ & \sc AUC $\uparrow$\\
\hline\hline
 \sc B1 & 2.3 \tiny (0.1) & 11.56 \tiny (0.2) & 12.69 \tiny(0.7) & 50.00 \tiny(0.0) \\ 
 \hline
 \sc B2 & 4.6 \tiny (0.0) & 13.22 \tiny(0.1) & 13.39 \tiny(0.8) & 50.01 \tiny(1.6)\\ 
 \hline
 \sc LR & 2.1 \tiny(0.1) & 11.0 \tiny(0.3) & 30.11 \tiny(1.2) & 71.67 \tiny(0.8)\\
 \hline
 \sc SVR & 2.3 \tiny(0.1) & 12.8 \tiny(0.1) & 28.66 \tiny(1.7) & 71.56 \tiny(1.0)\\
 \hline
 \sc RFs &  2.2 \tiny(0.1) & 11.2 \tiny(0.2) & 21.57 \tiny(1.0) & 59.67 \tiny(0.3)\\
 \hline
 \sc BERTr & \bf 1.8 \tiny(0.1) & \bf 9.2 \tiny(0.3)  &  \bf 42.01 \tiny(4.3) &  \bf 80.46 \tiny(1.3)\\
 \hline
 \end{tabular}
\caption{Mean Squared Error (MSE), Mean Absolute Error (MAE), Area Under Precision-Recall curve (AUPR), and ROC AUC of all \emph{context sensitivity estimation} models. An average (B1) and a random (B2) baseline have been included. All results averaged over three random splits, standard error of mean in brackets.}
\label{table:context_sensitivity_exps}
\end{table}

We used a train/validation/test split of 80/10/10 percent, respectively, and 
performed Monte Carlo 3-fold Cross Validation. We used mean square error (MSE) as our loss function and early stopping with patience of 5 epochs. Table~\ref{table:context_sensitivity_exps} presents the MSE and 
MAE of all the models on the test set. Unsurprisingly, BERTr outperforms the rest of the models in MSE and MAE. Previous work \cite{10.1145/3038912.3052591} reported that training toxicity regressors (based on the empirical distribution of codes) instead of classifiers (based on the majority of the codes) leads to improved classification results too, so we also computed classification results. For the latter results, we turned the ground truth probabilities of the test instances to binary labels by setting a threshold $t$ (Section~\ref{The_dataset}) and assigning the label 1 if $δ>t$ and 0 otherwise. In this experiment, $t$ was set to the sum of the standard error of mean (\textsc{sem}) of the \oc and \ic raters for that specific post $p$, i.e., $t(p) = \textsc{sem}^{oc}(p)+\textsc{sem}^{ic}(p)$. By using this binary ground truth, AUPR and AUC (Table~\ref{table:context_sensitivity_exps}) verified that BERTr outperforms the other models, even when the models are used as classifiers.

\section{Collecting Context Sensitive Posts}
\label{Context_sensitive_posts_retrieval}

In Section \ref{The_dataset} we saw that context sensitive posts
can be very
rare in toxicity datasets (Figure \ref{fig:context_sensitive_posts_per_delta}). In Section \ref{Experimental_Study} we 
showed that adding a simple context aware mechanism (concatenating the parent post) to 
an existing toxicity detection system can reduce the system’s error on context sensitive posts (Figure \ref{fig:Perspective_exps}). However, the error of the toxicity detector remains high for context-sensitive posts. This problem can potentially be addressed by augmenting the current datasets with more context-sensitive posts. As shown in Section \ref{Experimental_Study}, a regressor trained to predict the context sensitivity of a post can achieve low error (Table \ref{table:context_sensitivity_exps}). Hence, we assessed the scenario where 
a context sensitivity regressor is employed to obtain a dataset richer in context-sensitive posts.

\begin{figure}[t]
\includegraphics[width=0.45\textwidth]{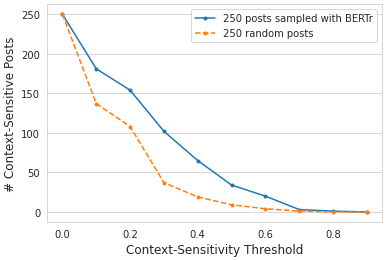}
\caption{Number of context-sensitive posts ($|\delta| \geq t$), for different context-sensitivity thresholds ($t$), using 250 likely context-sensitive posts sampled with BERTr (solid) or 250 randomly selected posts (dashed line).}
\label{fig:context_sensitive_posts_per_delta_250}
\end{figure}

We used our best context-sensitivity regressor (BERTr) to retrieve the 250 most likely context-sensitive posts from the 2M CC posts, excluding the 10,000 CCC posts. We then crowd-annotated the 250 posts in context (IC) as with CCC posts, keeping also the original out-of-context (OC) annotations. Table \ref{table:sampled_examples} shows examples of the 250 target posts obtained, along with their parent posts and δ. We then repeated the same experiment, this time using 250 \emph{randomly} selected posts from the 2M CC posts, excluding the 10,000 CCC posts and the 250 posts that were selected using BERTr. Figure~\ref{fig:context_sensitive_posts_per_delta_250} is the same as Figure~\ref{fig:context_sensitive_posts_per_delta}, but we now consider the 250 randomly selected posts (dashed line) and the 250 posts that were selected using BERTr (solid line). As in Figure~\ref{fig:context_sensitive_posts_per_delta}, we vary the context-sensitivity threshold $t$ on the horizontal axis. The 250 posts that were sampled using BERTr clearly include more context-sensitive posts than the 250 random ones, with the threshold ($t$) in the range $0.1 \leq t < 0.7$, indicating that BERTr can be successfully used to obtain datasets richer in context-sensitive posts. As in Figure~\ref{fig:context_sensitive_posts_per_delta}, there are very few context-sensitive posts for $t \geq 0.7$.

\begin{figure}[ht]
\centering
\includegraphics[width=0.48\textwidth]{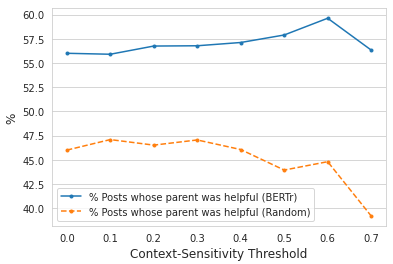}
\vspace*{-4mm}
\caption{Percentage of the 250 target posts, sampled with BERTr (solid) or random (dashed line), for which the majority of annotators found the parent post useful when assessing the toxicity of the target post.}
\label{fig:parent_utility}
\end{figure}

In this experiment, we also asked the crowd-annotators to indicate whether the parent post was helpful or not, when assessing the toxicity of each target post. Figure~\ref{fig:parent_utility} shows for how many of the 250 target posts (sampled using BERTr or random) the majority of the annotators responded that the parent post was useful. We vary the sensitivity threshold ($t$) on the horizontal axis up to $t=0.7$, since no posts are context-sensitive for $t > 0.7$ (Fig.~\ref{fig:context_sensitive_posts_per_delta_250}). The perceived utility of the parent posts is clearly higher for the 250 posts sampled with BERTr, compared to the 250 random ones, for all sensitivity thresholds. This again indicates that BERTr can be used to obtain datasets richer in context-sensitive posts.

\begin{table*}[h]
{\tiny
\centering
\begin{tabular}{ |p{5.5cm}|p{5.5cm}|c|c|c|c| } 
\hline
 \sc parent of post $p$ & \sc post $p$ &  $s^{OC}(p)$ & $s^{IC}(p)$ & $\delta$\\
\hline
 And since Thomas Aquinas never observed animals having gay sex in the wild homosexuality never made it into the annals of natural law theory. & Animals having "gay' sex? You mean there are "gay" animals.  So, when they're not "doing it" they do other things like go to Madonna concerts? & 60\% & 0\% & 60\%\\ 
 \hline
 Making a cake is MUCH different then selling gasoline or renting hotel rooms.  Making a cake is a form of artistry and requires the cake maker to artistically express him/herself which means the cake maker is actively participating.  Owning a gas station where random people pump their own gas does not require active participation. & Oh, ok. So the if the gas guy had to pump gas for that gay man, he should be able to refuse that, right? & 60\% & 0\% & 60\% \\ 
 \hline
  And SCOTUS will slap Watson \& Chinp down yet again....these Odummy Sock-Puppets never learn.  That threesome they shared back in the day must have been amazing. & Is the post implying that the judge is gay? I don't understand the comment, please explain? Are gays involved in this and not Muslims and their relatives? * & 83.3\% & 20\% & 63.3\% \\ 
 \hline
The appeal courts have one thing to do, ask is it legal or not, thats it, that is what appeals judges do, and they didnt, they coward away cause they knew then could not rule it illegal. sorry for your ignorance & The case has not yet been adjudicated on its merits (whether the Executive Order is illegal or not).  Both the trial decision and the appeal decision were about staying  the EO *until the trial on its merits* - ie, an injuction.  I'd think about finding out some facts before calling someone else ignorant, Rex. & 80\% & 20\% & 60\% \\ 
 \hline
 "...."marriage," by definition, meant one man, one woman ..."

Actually no. The definition restricting it to one man one woman unions was only introduced into USA law 2004/5/6 across numerous states in a frantic attempt to avoid courts making similar findings to those of the Massachusetts Supreme court ruling.
Prior to that it had always been expressly defined as between "two people", which is what triggered the Massachusets challenge.

The fact that only opposite sex marriages were performed in the past, does not mean marriage was defined as only between opposite sex couples, it simply illustrates that couples who were not opposite sex were being denied a fundamental right.

The evidence of the existence of discrimination is not proof that the discrimination was justified or justifiable.
& The definitive dictionary of the English language, the OED, does not contain a single instance in which "modern" civilized society has included gay marriage.  It does mention instances of "group" marriage in small, primitive societies, where all the men in a village are married to all the women.  But those, as you know, are by far the exception.  

Actually, your argument bolsters my point.  It was so universally understood at the founding of the Nation that marriage meant man-woman that marriage did not need to be defined.  Indeed, in most States, marriage could not have been defined so as to allow gay marriage, because until 1961, ALL 50 STATES outlawed sahdemy.  Do you begin to get at least part of the point?
 & 0\% & 60\% & -60\% \\ 
 \hline
 May be Trudeau should do a double apology just to one up Harper and then apologize for Papa Trudeau and no himself ruining the Canadian economy.& What has this got to do with the rape and abuse of boys and girls in residential schools? & 30\% & 60\% & -30\% \\ 
 \hline

\end{tabular}
}
\caption{Examples of context-sensitive posts in the sampled dataset. Here $s^{\oc}(p)$ and $s^{\ic}(p)$ are the fractions of out-of-context or in-context annotators, respectively, who found the target post $p$ to be toxic; and $\delta = s^{\oc}(p)-s^{\ic}(p)$.}
\label{table:sampled_examples}
\end{table*}

To verify the statistical significance of the finding that the annotators find the parent post useful more frequently in posts sampled with BERTr than in random posts, we performed a paired bootstrap resampling, following the experimental setting of \citet{Koehn2004}. We sampled 100 posts from the 250 random posts, and 100 posts from the 250 posts obtained by using BERTr, and we computed the percentage of posts where the majority of annotators found the parent post helpful, for random posts and BERTr posts. By resampling 1,000 times, we find that this percentage is greater for BERTr posts than for random posts, with a $P$-value of 0.05.

Finally, by turning the ground truth toxicity probabilities (for IC and OC annotation) into binary labels as in Section~\ref{Experimental_Study}, we estimated a context sensitivity class ratio (fraction of context-sensitive posts out of all 250 posts), for the BERTr-sampled and the randomly sampled posts. By using this class ratio, we found that 99 out of the 250 BERTr-sampled posts (39.6 percent) were context sensitive, while only 43 out of the 250 randomly sampled posts (17.2 percent) were context-sensitive (22 percent points lower; i.e., 57 percent decrease). We verified the statistical significance of this finding (lower fraction) by using bootstrapping with a $P$-value of 0.05, as in the previous paragraph. 
We conclude that sampling with BERTr leads to a higher context-sensitivity class ratio than random sampling. 

\section{Improving the Context-Sensitivity Regressor with Data Augmentation}

We showed in the previous section that by employing a context sensitivity regressor (BERTr was our best one) one can sample new sets of posts (e.g., from the 2M CC posts) that are richer in context-sensitive posts (by 22 percent points in our previous experiments) compared to random samples. By adding such richer (in context-sensitive posts) sets to an existing context sensitivity dataset (e.g., our CCC dataset), one can gradually increase the ratio of context-sensitive posts (which is low in CCC, see Fig.~\ref{fig:context_sensitive_posts_per_delta}).
A natural question then is whether one could improve the  context-sensitivity regressor by re-training it on the augmented dataset, which is less dominated by context-insensitive posts (more balanced in terms of context-sensitivity). Ideally the newly sampled (and overall more context-sensitive) posts would be crowd-annotated for context-sensitivity (by IC and OC raters) to obtain ground truth (gold context-sensitivity scores). To avoid this additional annotation cost, however, in this section we explore a teacher-student approach \cite{hinton2015distilling}. The teacher is the initial BERTr context-sensitivity regressor (Section \ref{context_sensitivity_experiments}), which provides silver context-sensitivity scores for the newly sampled posts. The student is another BERTr instance, which is trained on the augmented dataset (the data with gold sensitivity scores the teacher was trained on, plus the newly sampled posts with silver sensitivity scores). 
 
These steps can be repeated in cycles, by using the student as the new teacher to sample and silver-score additional posts in each cycle. Similar teacher-student approaches have recently been used in several NLP and computer vision tasks \cite{xie2020unsupervised, wei2018fast, Xie_2020_CVPR}, often using teacher and student models with different capacities. In our case, the teacher and student are the same model, but the student is trained on additional data silver-scored by the teacher, which is very similar to classical semi-supervised learning with Expectation Maximization \cite{bishop2006}.

Following this teacher-student approach, we experimented with data augmentation to improve the context-sensitivity estimator, using two different settings.
In both settings, the teacher silver-scores the newly sampled additional training posts. In the setting discussed first, the teacher is also used to sample the new training posts. By contrast, in the second setting the new posts are randomly sampled, and the teacher is only used to silver-score them.

\begin{figure}[h]
\centering
\includegraphics[width=0.48\textwidth]{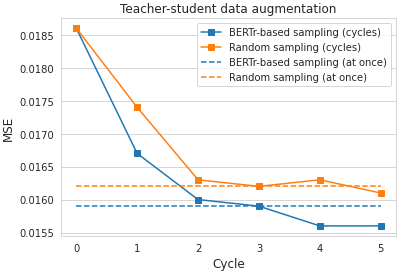}
\vspace*{-4mm}
\caption{Data augmentation with knowledge distillation to improve BERTr context-sensitivity regressor. 
Βlue solid line: the teacher model is used both to silver-score the new training instances and to sample them. Orange solid line: the teacher model is used only to silver-score the new training instances, which are randomly selected. Dashed lines: same as the solid ones, but only one cycle is performed, which adds 5,000 silver-scored new training instances at once. 
}
\label{fig:bootstrapping_MSE}
\end{figure}

\paragraph{Teacher-student with teacher sampling:} In this setting, we randomly sampled 20,000 posts from the 
Civil Comments (CC) dataset and used them as 
a pool to select (and silver-score) new training instances from, 
as follows:
\begin{enumerate}
  \item Train a BERTr teacher on the gold-scored (by crowd-annotators) training instances of our CCC dataset (Section~\ref{sec:CCC}).

  \item Use the BERTr teacher to silver-score for context-sensitivity  all the posts of the pool (initially 20,000).
  \item Select from the pool the 1,000 posts with the highest silver sensitivity scores, remove them from the pool, and add them (with their silver sensitivity scores) to the training set. 
  \item Train a BERTr student on the new training set (augmented by 1,000 silver-scored posts).
  \item Evaluate the student using exactly the same splits as in Section~\ref{context_sensitivity_experiments}.
  \item (Optional) Go back to step 2, using the student as a new teacher in a new cycle.
\end{enumerate}
We repeated this process for five cycles and ended up with a training set augmented by 5,000 likely context-sensitive posts. Experimental results (Fig.~\ref{fig:bootstrapping_MSE}, blue solid line) show performance gains in MSE even from the first cycle. We also compared against using a single cycle with 5,000 new posts added at once (blue dashed line), instead of adding only 1,000 posts per cycle and re-training the teacher. Performing cycles and re-training the teacher clearly leads to lower MSE, but with diminishing returns after cycle 4.

\paragraph{Teacher-student with random sampling:} This setting is the same as the previous one, but in step 3 we randomly select 1,000 posts from the pool, instead of selecting the 1,000 posts with the highest silver sensitivity scores. Again, we used five cycles (Fig.~\ref{fig:bootstrapping_MSE}, orange solid line) and we also compared to a single cycle that adds 5,000 silver-scored training instances at once (orange dashed line). Sampling with the teacher's scores (blue solid line) is clearly better than random sampling (orange lines).

\section{Related Work}

We describe related work along three dimensions. First, we describe work regarding toxicity detection. Second, we focus on context-aware natural language processing approaches. Third, we describe work that tackles classification tasks with regression-based approaches.

\subsection*{Toxicity detection}
Abusive language detection is a difficult task due to its subjective nature. Cyberbullies attack victims on different topics such as race, religion, and gender across multiple social media platforms \cite{agrawal}. Thus, the vocabulary used and the perceived meaning of words may vary when abusive language occurs in a different context. Several approaches have been examined in order to tackle the problem of abusive language detection. Researchers initially experimented with machine learning techniques using hand crafted features, such as lexical features, syntactic features, etc. \cite{Davidson_Warmsley_Macy_Weber_2017,waseem-hovy-2016-hateful,10.1145/2740908.2742760}. Then, deep learning techniques were employed, operating on word embeddings \cite{park-fung-2017-one, pavlopoulos-etal-2017-deeper,pavlopoulos-etal-2017-improved, chakrabarty-etal-2019-pay,Badjatiya, haddad-etal-2020-arabic, ozler-etal-2020-fine}. These techniques seem to work better for this task than the traditional machine learning methods based on handcrafted features \cite{Badjatiya}.

To facilitate research in this field, researchers have published several datasets containing different types of toxicity.
\citet{Nobata} developed a corpus of user comments posted on Yahoo Finance and News annotated for abusive language, the first of its kind. \citet{10.1145/3038912.3052591} created and experimented with three new datasets; the “Personal Attack” dataset where 115k comments from Wikipedia Talk pages were annotated as containing personal attack or not, the “Aggression” dataset where the same comments were annotated as being aggressive or not, and the “Toxicity” dataset that includes 159k comments again from Wikipedia Talk pages that were annotated as being toxic or not. \citet{waseem-hovy-2016-hateful} experimented on hate speech detection using a corpus of more than 16k tweets containing sexist, racist and non-toxic posts that they annotated by themselves. Most of the published toxicity datasets contain posts in English, but datasets in other languages also exist, such as Greek \cite{pavlopoulos-etal-2017-deep}, Arabic \cite{mubarak-etal-2017-abusive}, French \cite{chiril-etal-2020-annotated}, Indonesian \cite{IBROHIM2018222} and German \cite{Ross, Wiegand}.

\subsection*{Context-aware NLP}
Incorporating context into human language technology has been successfully applied to various applications and domains. In text/word representation, context has a central role \cite{Mikolov,pennington-etal-2014-glove, melamud-etal-2016-context2vec, peters-etal-2018-deep, devlin-etal-2019-bert}. Integrating context is crucial in the sentiment analysis task too, where the semantic orientation of a word changes according to the domain or the context in which that word is being used \cite{Agarwal-and-Basant}. \citet{vanzo-etal-2014-context} explored the role of contextual information in supervised sentiment analysis over Twitter. They proposed two different types of contexts, a conversation-based context and a topic-based context, which includes several tweets in the history stream that contain overlapping hashtags. They modeled each tweet and its context as a sequence of tweets and used a sequence labeling model, SVM$^{\mbox{\scriptsize \it HMM}}$, to predict their sentiment labels jointly. They found that the kind of context they considered leads to specific consistent benefits in sentiment classification. \citet{10.5555/3015812.3015844} proposed a context-based neural network model for Twitter sentiment analysis, incorporating contextualized features from relevant tweets into the model in the form of word embedding vectors. They experimented with three types of context, a conversation-based context, an author-based context, and a topic-based context. They found that integrating contextual information about the target tweet in their neural model offers improved performance compared with the state-of-the-art discrete and continuous word representation models. They also reported that topic-based context features  were the most effective for this task.

Despite the wide use of context in other Natural Language Processing (NLP) tasks, such as dialogue systems \cite{Lowe2015IncorporatingUT, dusek-jurcicek-2016-context} and informational bias detection \cite{van-den-berg-markert-2020-context}, very 
few researchers have focused on context-aware toxic language detection. \citet{gao-huang-2017-detecting} provided a corpus of speech labeled by annotators as hateful, obtained from full threads of online discussion posts under Fox News articles. They proposed two types of hate speech detection models that incorporate context information, a logistic regression model with context features and a neural network model with learning components for context. They reported performance gains in F1 score when incorporating context and that combining the two types of models they considered further improved performance by another 7 percent in F1 score. \citet{mubarak-etal-2017-abusive} provided the title of the respective news article to the annotators during the annotation process, but they ignored parent comments since they did not have the entire thread. As \citet{pavlopoulos2020toxicity} already noticed, this presents the following problem: new comments may change the topic of the conversation and  replies may require the previous posts to be assessed correctly. \citet{pavlopoulos-etal-2017-deep} provided the annotators with the whole conversation thread for each target comment as context during the annotation process. The plain text of the comments is not available, however, which makes further analysis difficult. 

In later work \citet{pavlopoulos2020toxicity} published two new toxicity datasets containing posts from the Wikipedia Talk pages, where during the annotation process, annotators were provided with the previous post in the thread and the discussion title. The authors found that providing annotators with  context can result both in amplification or mitigation of the perceived toxicity of posts. Moreover, they found no evidence that context actually improves the performance of toxicity classifiers. In a similar work that was conducted by \citet{menini2021abuse}, the authors investigated the role of textual context in abusive language detection on Twitter. They first re-annotated the tweets in the dataset of \citet{founta2018large} in two settings, with and without context. After comparing the two datasets (with and without context-aware annotations) they found that context is sometimes necessary to understand the real intent of the user, and that it is more likely to mitigate the abusiveness of a tweet even if it contains profanity. Finally, they experimented with several classifiers, using both context-aware and context-unaware architectures. Their experimental results showed that when classifiers are given context and are evaluated on context-aware
datasets, their performance drops dramatically compared to a setting where classifiers are not given context and are evaluated on context-unaware datasets.

\subsection*{Regression as classification in NLP}
Approaching a text classification problem as a regression-based problem has been tested by researchers in various NLP tasks, such as sentiment analysis \cite{wang-etal-2016-dimensional}, emotional analysis \cite{10.3233/978-1-61499-672-9-1114}, metaphor detection \cite{Parde_Nielsen_2018}, toxicity detection. A work similar to ours in that respect is the work of  \citet{10.1145/3038912.3052591}, 
who noticed that estimating the likelihood of a post to be personal attack or not, using the empirical distribution of human-ratings, rather than the majority vote, produces a better classifier, even in terms of the ROC AUC metric. \citet{dsa-etal-2020-towards} experimented on the English Wikipedia Detox corpus by designing both binary classification and regression-based approaches aiming to predict whether a comment is toxic or not. They compared different unsupervised word representations and different deep learning based classifiers. In most of their experiments, the regression-based approach showed slightly better performance than the classification setting, which is consistent with the findings of \citet{10.1145/3038912.3052591}.
 
\section{Limitations and Considerations}
\begin{itemize}
    \item We limited our study to the parent post of the conversational context, but we note that more posts or even the entire thread could be used. Also, other possible sources of context exist and could be examined along the thread’s posts. For instance, the discussion title could provide the annotators with more information when they annotate each post of the discussion. We consider this study as the first of a series of steps that need to be taken to investigate the relation of context in toxicity detection.   
    \item Online discussions are currently moderated by human raters and machine learning models. 
    Both may carry bias introduced by the annotators (e.g., if all the annotators originate from the same cultural background). The same limitation applies to this study, for which we employed crowd annotators, but without trying to control for possibly different social norms.
    \item We focused on English posts and we employed English-speaking annotators. The English-centric nature of the Internet is a widely acknowledged problem. The ways that the communicative intent is mixed into culture, and the notions of what is appropriate in a given context are problems that are sometimes simpler for a language like English that does not mark gender and has lost its formal/casual distinctions. This is also related to the difficulty of doing work with social norms, which is challenging in well resourced languages and virtually impossible in languages that lack modeling resources.
\end{itemize}
The above mentioned limitations only highlight the challenges of the work that remains to be done.

\section{Conclusions and Future Work}
We introduced the task of estimating the context-sensitivity of posts in toxicity detection, i.e., estimating the extent to which the perceived toxicity of a post  depends on the conversational context. We constructed, presented, and released a new dataset that can be used to train and evaluate systems for the new task, where context is the previous post. We also showed that context-sensitivity estimation systems can be used to collect larger samples of context-sensitive posts, which is a prerequisite to train toxicity detectors to better handle context-sensitive posts. Furthermore, we showed that the performance or our best context sensitivity estimator is further improved by augmenting the training dataset with teacher-student knowledge distillation. Context-sensitivity estimators can also be used to suggest when moderators should consider the context of a post, which is more costly and may not always be necessary.  
In future work, we hope to incorporate context mechanisms in toxicity detectors and train (and evaluate) them on datasets sufficiently rich in context-sensitive posts.

\bibliographystyle{acl_natbib}
\bibliography{acl2021}

\begin{thebibliography}{50}
\expandafter\ifx\csname natexlab\endcsname\relax\def\natexlab#1{#1}\fi

\bibitem[{Agarwal et~al.(2015)Agarwal, Mittal, Bansal, and
  Garg}]{Agarwal-and-Basant}
Basant Agarwal, Namita Mittal, Pooja Bansal, and Sonal Garg. 2015.
\newblock \href {https://doi.org/10.1155/2015/715730} {Sentiment analysis using
  common-sense and context information}.
\newblock \emph{Computational intelligence and neuroscience}, 2015:715730.

\bibitem[{Agrawal and Awekar(2018)}]{agrawal}
Sweta Agrawal and Amit Awekar. 2018.
\newblock Deep learning for detecting cyberbullying across multiple social
  media platforms.
\newblock \emph{ArXiv}, abs/1801.06482.

\bibitem[{Badjatiya et~al.(2017)Badjatiya, Gupta, Gupta, and Varma}]{Badjatiya}
Pinkesh Badjatiya, Shashank Gupta, Manish Gupta, and Vasudeva Varma. 2017.
\newblock \href {https://doi.org/10.1145/3041021.3054223} {Deep learning for
  hate speech detection in tweets}.
\newblock \emph{Proceedings of the 26th International Conference on World Wide
  Web Companion - WWW ’17 Companion}.

\bibitem[{van~den Berg and Markert(2020)}]{van-den-berg-markert-2020-context}
Esther van~den Berg and Katja Markert. 2020.
\newblock \href {https://doi.org/10.18653/v1/2020.coling-main.556} {Context in
  informational bias detection}.
\newblock In \emph{Proceedings of the 28th International Conference on
  Computational Linguistics}, pages 6315--6326, Barcelona, Spain (Online).
  International Committee on Computational Linguistics.

\bibitem[{Bishop(2006)}]{bishop2006}
Christopher~M Bishop. 2006.
\newblock Pattern recognition.
\newblock \emph{Machine learning}, 128(9).

\bibitem[{Borkan et~al.(2019)Borkan, Dixon, Sorensen, Thain, and
  Vasserman}]{Borkan2019}
Daniel Borkan, Lucas Dixon, Jeffrey Sorensen, Nithum Thain, and Lucy Vasserman.
  2019.
\newblock Nuanced metrics for measuring unintended bias with real data for text
  classification.
\newblock In \emph{WWW}, pages 491--500, San Francisco, USA.

\bibitem[{Buechel and Hahn(2016)}]{10.3233/978-1-61499-672-9-1114}
Sven Buechel and Udo Hahn. 2016.
\newblock \href {https://doi.org/10.3233/978-1-61499-672-9-1114} {Emotion
  analysis as a regression problem — dimensional models and their
  implications on emotion representation and metrical evaluation}.
\newblock In \emph{Proceedings of the Twenty-Second European Conference on
  Artificial Intelligence}, ECAI'16, page 1114–1122, NLD. IOS Press.

\bibitem[{Chakrabarty et~al.(2019)Chakrabarty, Gupta, and
  Muresan}]{chakrabarty-etal-2019-pay}
Tuhin Chakrabarty, Kilol Gupta, and Smaranda Muresan. 2019.
\newblock \href {https://doi.org/10.18653/v1/W19-3508} {Pay {``}attention{''}
  to your context when classifying abusive language}.
\newblock In \emph{Proceedings of the Third Workshop on Abusive Language
  Online}, pages 70--79, Florence, Italy. Association for Computational
  Linguistics.

\bibitem[{Chiril et~al.(2020)Chiril, Moriceau, Benamara, Mari, Origgi, and
  Coulomb-Gully}]{chiril-etal-2020-annotated}
Patricia Chiril, V{\'e}ronique Moriceau, Farah Benamara, Alda Mari, Gloria
  Origgi, and Marl{\`e}ne Coulomb-Gully. 2020.
\newblock \href {https://www.aclweb.org/anthology/2020.lrec-1.175} {An
  annotated corpus for sexism detection in {F}rench tweets}.
\newblock In \emph{Proceedings of the 12th Language Resources and Evaluation
  Conference}, pages 1397--1403, Marseille, France. European Language Resources
  Association.

\bibitem[{Davidson et~al.(2017)Davidson, Warmsley, Macy, and
  Weber}]{Davidson_Warmsley_Macy_Weber_2017}
Thomas Davidson, Dana Warmsley, Michael Macy, and Ingmar Weber. 2017.
\newblock \href {https://ojs.aaai.org/index.php/ICWSM/article/view/14955}
  {Automated hate speech detection and the problem of offensive language}.
\newblock \emph{Proceedings of the International AAAI Conference on Web and
  Social Media}, 11(1).

\bibitem[{Devlin et~al.(2019)Devlin, Chang, Lee, and
  Toutanova}]{devlin-etal-2019-bert}
Jacob Devlin, Ming-Wei Chang, Kenton Lee, and Kristina Toutanova. 2019.
\newblock \href {https://doi.org/10.18653/v1/N19-1423} {{BERT}: Pre-training of
  deep bidirectional transformers for language understanding}.
\newblock In \emph{Proceedings of the 2019 Conference of the North {A}merican
  Chapter of the Association for Computational Linguistics: Human Language
  Technologies, Volume 1 (Long and Short Papers)}, pages 4171--4186,
  Minneapolis, Minnesota. Association for Computational Linguistics.

\bibitem[{Djuric et~al.(2015)Djuric, Zhou, Morris, Grbovic, Radosavljevic, and
  Bhamidipati}]{10.1145/2740908.2742760}
Nemanja Djuric, Jing Zhou, Robin Morris, Mihajlo Grbovic, Vladan Radosavljevic,
  and Narayan Bhamidipati. 2015.
\newblock \href {https://doi.org/10.1145/2740908.2742760} {Hate speech
  detection with comment embeddings}.
\newblock In \emph{Proceedings of the 24th International Conference on World
  Wide Web}, WWW '15 Companion, page 29–30, New York, NY, USA. Association
  for Computing Machinery.

\bibitem[{D{'}Sa et~al.(2020)D{'}Sa, Illina, and Fohr}]{dsa-etal-2020-towards}
Ashwin~Geet D{'}Sa, Irina Illina, and Dominique Fohr. 2020.
\newblock \href {https://www.aclweb.org/anthology/2020.trac-1.4} {Towards
  non-toxic landscapes: Automatic toxic comment detection using {DNN}}.
\newblock In \emph{Proceedings of the Second Workshop on Trolling, Aggression
  and Cyberbullying}, pages 21--25, Marseille, France. European Language
  Resources Association (ELRA).

\bibitem[{Du{\v{s}}ek and
  Jur{\v{c}}{\'\i}{\v{c}}ek(2016)}]{dusek-jurcicek-2016-context}
Ond{\v{r}}ej Du{\v{s}}ek and Filip Jur{\v{c}}{\'\i}{\v{c}}ek. 2016.
\newblock \href {https://doi.org/10.18653/v1/W16-3622} {A context-aware natural
  language generator for dialogue systems}.
\newblock In \emph{Proceedings of the 17th Annual Meeting of the Special
  Interest Group on Discourse and Dialogue}, pages 185--190, Los Angeles.
  Association for Computational Linguistics.

\bibitem[{Fornaciari et~al.(2021)Fornaciari, Uma, Paun, Plank, Hovy, and
  Poesio}]{fornaciari-etal-2021-beyond}
Tommaso Fornaciari, Alexandra Uma, Silviu Paun, Barbara Plank, Dirk Hovy, and
  Massimo Poesio. 2021.
\newblock \href {https://doi.org/10.18653/v1/2021.naacl-main.204} {Beyond black
  {\&} white: Leveraging annotator disagreement via soft-label multi-task
  learning}.
\newblock In \emph{Proceedings of the 2021 Conference of the North American
  Chapter of the Association for Computational Linguistics: Human Language
  Technologies}, pages 2591--2597, Online. Association for Computational
  Linguistics.

\bibitem[{Founta et~al.(2018)Founta, Djouvas, Chatzakou, Leontiadis, Blackburn,
  Stringhini, Vakali, Sirivianos, and Kourtellis}]{founta2018large}
Antigoni-Maria Founta, Constantinos Djouvas, Despoina Chatzakou, Ilias
  Leontiadis, Jeremy Blackburn, Gianluca Stringhini, Athena Vakali, Michael
  Sirivianos, and Nicolas Kourtellis. 2018.
\newblock \href {http://arxiv.org/abs/1802.00393} {Large scale crowdsourcing
  and characterization of twitter abusive behavior}.

\bibitem[{Gao and Huang(2017)}]{gao-huang-2017-detecting}
Lei Gao and Ruihong Huang. 2017.
\newblock \href {https://doi.org/10.26615/978-954-452-049-6_036} {Detecting
  online hate speech using context aware models}.
\newblock In \emph{Proceedings of the International Conference Recent Advances
  in Natural Language Processing, {RANLP} 2017}, pages 260--266, Varna,
  Bulgaria. INCOMA Ltd.

\bibitem[{Haddad et~al.(2020)Haddad, Orabe, Al-Abood, and
  Ghneim}]{haddad-etal-2020-arabic}
Bushr Haddad, Zoher Orabe, Anas Al-Abood, and Nada Ghneim. 2020.
\newblock \href {https://www.aclweb.org/anthology/2020.osact-1.12} {{A}rabic
  offensive language detection with attention-based deep neural networks}.
\newblock In \emph{Proceedings of the 4th Workshop on Open-Source Arabic
  Corpora and Processing Tools, with a Shared Task on Offensive Language
  Detection}, pages 76--81, Marseille, France. European Language Resource
  Association.

\bibitem[{Hinton et~al.(2015)Hinton, Vinyals, and Dean}]{hinton2015distilling}
Geoffrey Hinton, Oriol Vinyals, and Jeff Dean. 2015.
\newblock \href {http://arxiv.org/abs/1503.02531} {Distilling the knowledge in
  a neural network}.

\bibitem[{Ibrohim and Budi(2018)}]{IBROHIM2018222}
Muhammad~Okky Ibrohim and Indra Budi. 2018.
\newblock \href {https://doi.org/https://doi.org/10.1016/j.procs.2018.08.169}
  {A dataset and preliminaries study for abusive language detection in
  indonesian social media}.
\newblock \emph{Procedia Computer Science}, 135:222--229.
\newblock The 3rd International Conference on Computer Science and
  Computational Intelligence (ICCSCI 2018) : Empowering Smart Technology in
  Digital Era for a Better Life.

\bibitem[{Koehn(2004)}]{Koehn2004}
Philipp Koehn. 2004.
\newblock Statistical significance tests for machine translation evaluation.
\newblock In \emph{Proceedings of the 2004 conference on empirical methods in
  natural language processing}, pages 388--395.

\bibitem[{Lowe et~al.(2015)Lowe, Pow, Charlin, and
  Pineau}]{Lowe2015IncorporatingUT}
R.~Lowe, Nissan Pow, Laurent Charlin, and Joelle Pineau. 2015.
\newblock Incorporating unstructured textual knowledge sources into neural
  dialogue systems.
\newblock In \emph{Machine Learning for Spoken Language Understanding and
  Interaction, NIPS 2015 Workshop}.

\bibitem[{Melamud et~al.(2016)Melamud, Goldberger, and
  Dagan}]{melamud-etal-2016-context2vec}
Oren Melamud, Jacob Goldberger, and Ido Dagan. 2016.
\newblock \href {https://doi.org/10.18653/v1/K16-1006} {context2vec: Learning
  generic context embedding with bidirectional {LSTM}}.
\newblock In \emph{Proceedings of The 20th {SIGNLL} Conference on Computational
  Natural Language Learning}, pages 51--61, Berlin, Germany. Association for
  Computational Linguistics.

\bibitem[{Menini et~al.(2021)Menini, Aprosio, and Tonelli}]{menini2021abuse}
Stefano Menini, Alessio~Palmero Aprosio, and Sara Tonelli. 2021.
\newblock \href {http://arxiv.org/abs/2103.14916} {Abuse is contextual, what
  about nlp? the role of context in abusive language annotation and detection}.

\bibitem[{Mikolov et~al.(2013)Mikolov, Sutskever, Chen, Corrado, and
  Dean}]{Mikolov}
Tomas Mikolov, Ilya Sutskever, Kai Chen, Greg Corrado, and Jeffrey Dean. 2013.
\newblock Distributed representations of words and phrases and their
  compositionality.
\newblock In \emph{Proceedings of the 26th International Conference on Neural
  Information Processing Systems - Volume 2}, NIPS'13, page 3111–3119, Red
  Hook, NY, USA. Curran Associates Inc.

\bibitem[{Mubarak et~al.(2017)Mubarak, Darwish, and
  Magdy}]{mubarak-etal-2017-abusive}
Hamdy Mubarak, Kareem Darwish, and Walid Magdy. 2017.
\newblock \href {https://doi.org/10.18653/v1/W17-3008} {Abusive language
  detection on {A}rabic social media}.
\newblock In \emph{Proceedings of the First Workshop on Abusive Language
  Online}, pages 52--56, Vancouver, BC, Canada. Association for Computational
  Linguistics.

\bibitem[{Nobata et~al.(2016)Nobata, Tetreault, Thomas, Mehdad, and
  Chang}]{Nobata}
Chikashi Nobata, Joel Tetreault, Achint Thomas, Yashar Mehdad, and Yi~Chang.
  2016.
\newblock \href {https://doi.org/10.1145/2872427.2883062} {Abusive language
  detection in online user content}.
\newblock In \emph{Proceedings of the 25th International Conference on World
  Wide Web}, WWW '16, page 145–153, Republic and Canton of Geneva, CHE.
  International World Wide Web Conferences Steering Committee.

\bibitem[{Ozler et~al.(2020)Ozler, Kenski, Rains, Shmargad, Coe, and
  Bethard}]{ozler-etal-2020-fine}
Kadir~Bulut Ozler, Kate Kenski, Steve Rains, Yotam Shmargad, Kevin Coe, and
  Steven Bethard. 2020.
\newblock \href {https://doi.org/10.18653/v1/2020.alw-1.4} {Fine-tuning for
  multi-domain and multi-label uncivil language detection}.
\newblock In \emph{Proceedings of the Fourth Workshop on Online Abuse and
  Harms}, pages 28--33, Online. Association for Computational Linguistics.

\bibitem[{Parde and Nielsen(2018)}]{Parde_Nielsen_2018}
Natalie Parde and Rodney Nielsen. 2018.
\newblock \href {https://ojs.aaai.org/index.php/AAAI/article/view/11940}
  {Exploring the terrain of metaphor novelty: A regression-based approach for
  automatically scoring metaphors}.
\newblock \emph{Proceedings of the AAAI Conference on Artificial Intelligence},
  32(1).

\bibitem[{Park and Fung(2017)}]{park-fung-2017-one}
Ji~Ho Park and Pascale Fung. 2017.
\newblock \href {https://doi.org/10.18653/v1/W17-3006} {One-step and two-step
  classification for abusive language detection on {T}witter}.
\newblock In \emph{Proceedings of the First Workshop on Abusive Language
  Online}, pages 41--45, Vancouver, BC, Canada. Association for Computational
  Linguistics.

\bibitem[{Pavlopoulos et~al.(2017{\natexlab{a}})Pavlopoulos, Malakasiotis, and
  Androutsopoulos}]{pavlopoulos-etal-2017-deep}
John Pavlopoulos, Prodromos Malakasiotis, and Ion Androutsopoulos.
  2017{\natexlab{a}}.
\newblock \href {https://doi.org/10.18653/v1/W17-3004} {Deep learning for user
  comment moderation}.
\newblock In \emph{Proceedings of the First Workshop on Abusive Language
  Online}, pages 25--35, Vancouver, BC, Canada. Association for Computational
  Linguistics.

\bibitem[{Pavlopoulos et~al.(2017{\natexlab{b}})Pavlopoulos, Malakasiotis, and
  Androutsopoulos}]{pavlopoulos-etal-2017-deeper}
John Pavlopoulos, Prodromos Malakasiotis, and Ion Androutsopoulos.
  2017{\natexlab{b}}.
\newblock \href {https://doi.org/10.18653/v1/D17-1117} {Deeper attention to
  abusive user content moderation}.
\newblock In \emph{Proceedings of the 2017 Conference on Empirical Methods in
  Natural Language Processing}, pages 1125--1135, Copenhagen, Denmark.
  Association for Computational Linguistics.

\bibitem[{Pavlopoulos et~al.(2017{\natexlab{c}})Pavlopoulos, Malakasiotis,
  Bakagianni, and Androutsopoulos}]{pavlopoulos-etal-2017-improved}
John Pavlopoulos, Prodromos Malakasiotis, Juli Bakagianni, and Ion
  Androutsopoulos. 2017{\natexlab{c}}.
\newblock \href {https://doi.org/10.18653/v1/W17-4209} {Improved abusive
  comment moderation with user embeddings}.
\newblock In \emph{Proceedings of the 2017 {EMNLP} Workshop: Natural Language
  Processing meets Journalism}, pages 51--55, Copenhagen, Denmark. Association
  for Computational Linguistics.

\bibitem[{Pavlopoulos et~al.(2020)Pavlopoulos, Sorensen, Dixon, Thain, and
  Androutsopoulos}]{pavlopoulos2020toxicity}
John Pavlopoulos, Jeffrey Sorensen, Lucas Dixon, Nithum Thain, and Ion
  Androutsopoulos. 2020.
\newblock \href {http://arxiv.org/abs/2006.00998} {Toxicity detection: Does
  context really matter?}

\bibitem[{Pennington et~al.(2014)Pennington, Socher, and
  Manning}]{pennington-etal-2014-glove}
Jeffrey Pennington, Richard Socher, and Christopher Manning. 2014.
\newblock \href {https://doi.org/10.3115/v1/D14-1162} {{G}lo{V}e: Global
  vectors for word representation}.
\newblock In \emph{Proceedings of the 2014 Conference on Empirical Methods in
  Natural Language Processing ({EMNLP})}, pages 1532--1543, Doha, Qatar.
  Association for Computational Linguistics.

\bibitem[{Peters et~al.(2018)Peters, Neumann, Iyyer, Gardner, Clark, Lee, and
  Zettlemoyer}]{peters-etal-2018-deep}
Matthew Peters, Mark Neumann, Mohit Iyyer, Matt Gardner, Christopher Clark,
  Kenton Lee, and Luke Zettlemoyer. 2018.
\newblock \href {https://doi.org/10.18653/v1/N18-1202} {Deep contextualized
  word representations}.
\newblock In \emph{Proceedings of the 2018 Conference of the North {A}merican
  Chapter of the Association for Computational Linguistics: Human Language
  Technologies, Volume 1 (Long Papers)}, pages 2227--2237, New Orleans,
  Louisiana. Association for Computational Linguistics.

\bibitem[{Randolph(2010)}]{Randolph}
Justus Randolph. 2010.
\newblock Free-marginal multirater kappa (multirater κfree): An alternative to
  fleiss fixed-marginal multirater kappa.
\newblock volume~4.

\bibitem[{Ren et~al.(2016)Ren, Zhang, Zhang, and Ji}]{10.5555/3015812.3015844}
Yafeng Ren, Yue Zhang, Meishan Zhang, and Donghong Ji. 2016.
\newblock Context-sensitive twitter sentiment classification using neural
  network.
\newblock In \emph{Proceedings of the Thirtieth AAAI Conference on Artificial
  Intelligence}, AAAI'16, page 215–221. AAAI Press.

\bibitem[{Ross et~al.(2016)Ross, Rist, Carbonell, Cabrera, Kurowsky, and
  Wojatzki}]{Ross}
Bj{\"{o}}rn Ross, Michael Rist, Guillermo Carbonell, Ben Cabrera, Nils
  Kurowsky, and Michael Wojatzki. 2016.
\newblock \href {https://arxiv.org/pdf/1701.08118.pdf} {{Measuring the
  Reliability of Hate Speech Annotations: The Case of the European Refugee
  Crisis}}.
\newblock In \emph{Proceedings of NLP4CMC III: 3rd Workshop on Natural Language
  Processing for Computer-Mediated Communication}, pages 6--9.

\bibitem[{Sap et~al.(2020)Sap, Gabriel, Qin, Jurafsky, Smith, and
  Choi}]{Sap2020}
Maarten Sap, Saadia Gabriel, Lianhui Qin, Dan Jurafsky, Noah~A. Smith, and
  Yejin Choi. 2020.
\newblock \href {https://doi.org/10.18653/v1/2020.acl-main.486} {Social bias
  frames: Reasoning about social and power implications of language}.
\newblock In \emph{Proceedings of the 58th Annual Meeting of the Association
  for Computational Linguistics}, pages 5477--5490, Online. Association for
  Computational Linguistics.

\bibitem[{Thylstrup and Waseem(2020)}]{Thylstrup2020}
Nanna Thylstrup and Zeerak Waseem. 2020.
\newblock \href {https://doi.org/10.2139/ssrn.3709719} {Detecting ‘dirt’
  and ‘toxicity’: Rethinking content moderation as pollution behaviour}.
\newblock \emph{SSRN Electronic Journal}.

\bibitem[{Vanzo et~al.(2014)Vanzo, Croce, and Basili}]{vanzo-etal-2014-context}
Andrea Vanzo, Danilo Croce, and Roberto Basili. 2014.
\newblock \href {https://www.aclweb.org/anthology/C14-1221} {A context-based
  model for sentiment analysis in {T}witter}.
\newblock In \emph{Proceedings of {COLING} 2014, the 25th International
  Conference on Computational Linguistics: Technical Papers}, pages 2345--2354,
  Dublin, Ireland. Dublin City University and Association for Computational
  Linguistics.

\bibitem[{Wang et~al.(2016)Wang, Yu, Lai, and
  Zhang}]{wang-etal-2016-dimensional}
Jin Wang, Liang-Chih Yu, K.~Robert Lai, and Xuejie Zhang. 2016.
\newblock \href {https://doi.org/10.18653/v1/P16-2037} {Dimensional sentiment
  analysis using a regional {CNN}-{LSTM} model}.
\newblock In \emph{Proceedings of the 54th Annual Meeting of the Association
  for Computational Linguistics (Volume 2: Short Papers)}, pages 225--230,
  Berlin, Germany. Association for Computational Linguistics.

\bibitem[{Waseem and Hovy(2016)}]{waseem-hovy-2016-hateful}
Zeerak Waseem and Dirk Hovy. 2016.
\newblock \href {https://doi.org/10.18653/v1/N16-2013} {Hateful symbols or
  hateful people? predictive features for hate speech detection on {T}witter}.
\newblock In \emph{Proceedings of the {NAACL} Student Research Workshop}, pages
  88--93, San Diego, California. Association for Computational Linguistics.

\bibitem[{Wiegand et~al.(2018)Wiegand, Siegel, and Ruppenhofer}]{Wiegand}
Michael Wiegand, Melanie Siegel, and Josef Ruppenhofer. 2018.
\newblock Overview of the germeval 2018 shared task on the identification of
  offensive language.
\newblock In \emph{Proceedings of GermEval 2018, 14th Conference on Natural
  Language Processing (KONVENS 2018), Vienna, Austria – September 21, 2018}.

\bibitem[{Wulczyn et~al.(2017)Wulczyn, Thain, and
  Dixon}]{10.1145/3038912.3052591}
Ellery Wulczyn, Nithum Thain, and Lucas Dixon. 2017.
\newblock \href {https://doi.org/10.1145/3038912.3052591} {Ex machina: Personal
  attacks seen at scale}.
\newblock In \emph{Proceedings of the 26th International Conference on World
  Wide Web}, WWW '17, page 1391–1399, Republic and Canton of Geneva, CHE.
  International World Wide Web Conferences Steering Committee.

\bibitem[{Xie et~al.(2020{\natexlab{a}})Xie, Dai, Hovy, Luong, and
  Le}]{xie2020unsupervised}
Qizhe Xie, Zihang Dai, Eduard Hovy, Minh-Thang Luong, and Quoc~V. Le.
  2020{\natexlab{a}}.
\newblock \href {https://openreview.net/forum?id=ByeL1R4FvS} {Unsupervised data
  augmentation for consistency training}.

\bibitem[{Xie et~al.(2020{\natexlab{b}})Xie, Luong, Hovy, and
  Le}]{Xie_2020_CVPR}
Qizhe Xie, Minh-Thang Luong, Eduard Hovy, and Quoc~V. Le. 2020{\natexlab{b}}.
\newblock Self-training with noisy student improves imagenet classification.
\newblock In \emph{Proceedings of the IEEE/CVF Conference on Computer Vision
  and Pattern Recognition (CVPR)}.

\bibitem[{Yu et~al.(2018)Yu, Dohan, Le, Luong, Zhao, and Chen}]{wei2018fast}
Adams~Wei Yu, David Dohan, Quoc Le, Thang Luong, Rui Zhao, and Kai Chen. 2018.
\newblock \href {https://openreview.net/forum?id=B14TlG-RW} {Fast and accurate
  reading comprehension by combining self-attention and convolution}.
\newblock In \emph{International Conference on Learning Representations}.

\bibitem[{Zhang et~al.(2018)Zhang, Chang, Danescu-Niculescu-Mizil, Dixon, Hua,
  Taraborelli, and Thain}]{zhang-etal-2018-conversations}
Justine Zhang, Jonathan Chang, Cristian Danescu-Niculescu-Mizil, Lucas Dixon,
  Yiqing Hua, Dario Taraborelli, and Nithum Thain. 2018.
\newblock \href {https://doi.org/10.18653/v1/P18-1125} {Conversations gone
  awry: Detecting early signs of conversational failure}.
\newblock In \emph{Proceedings of the 56th Annual Meeting of the Association
  for Computational Linguistics (Volume 1: Long Papers)}, pages 1350--1361,
  Melbourne, Australia. Association for Computational Linguistics.

\end{thebibliography}

\end{document}